%% file: acl_latex.tex
\documentclass[11pt]{article}

\usepackage[preprint]{acl}

\usepackage{times}
\usepackage{latexsym}
\usepackage{listings}
\usepackage{marvosym}
\usepackage[T1]{fontenc}

\usepackage[utf8]{inputenc}

\usepackage{microtype}

\usepackage{inconsolata}

\usepackage{graphicx}

\usepackage{amsmath}
\usepackage{booktabs}
\usepackage{multirow}
\usepackage{pifont}
\newcommand{\cmark}{\ding{51}}
\newcommand{\xmark}{\ding{55}}
\usepackage{adjustbox}
\usepackage{tabularx}
\usepackage{amsmath,amssymb}
\usepackage{multicol}

\title{OmniInteract: Benchmarking Real-World Streaming Interaction for Real-Time Omnimodal Assistants}

\author{\bfseries
Xudong Lu$^{*1\dagger}$, Xueying Li$^{*2}$, Annan Wang$^{*3}$, Yang Bo$^{4}$, Jinpeng Chen$^{5}$, Zengliang Li$^{6}$,\\
\bfseries
Nianzu Yang$^{2}$, Rui Liu$^{1}$, Xue Yang$^{2}$, Jingwen Hou$^{6{\textrm{\Letter}}}$, Hongsheng Li$^{1}$\\
\text{$^1$CUHK MMLab\quad $^2$SJTU\quad $^3$NTU\quad $^4$McMaster\quad $^5$CityUHK\quad $^6$JUFE}\\
\texttt{luxudong@link.cuhk.edu.hk, jingwen003@e.ntu.edu.sg}\\
\small $^*$Equal contribution \quad $^{\textrm{\Letter}}$Corresponding author \quad $^\dagger$Project lead
}

\begin{document}
\maketitle

\input{src/0_abs}
\input{src/1_intro}

\input{src/2_related}
\input{src/3_method}
\input{src/4_exp}

\input{src/5_conclusion}

\bibliography{custom}

\input{src/x_appendix}

\end{document}

%% file: src/0_abs.tex
\begin{abstract}
We introduce OmniInteract, a streaming benchmark for real-time omnimodal large language models evaluated through native online inference over audio-visual streams. Unlike offline video understanding or text-prompted streaming QA, OmniInteract preserves the original audio-visual stream and requires models to process it online, without access to future content. User queries and ambient sounds are embedded in the audio track, requiring models to detect multimodal triggers, decide when to respond, and answer while the stream unfolds. OmniInteract contains 250 videos with 1,430 temporally grounded response slots: 1,062 1Q1A slots across real-time, proactive, and nested scenarios, and 368 1QnA slots for continuous task monitoring and step guidance. Each slot includes a trigger, response window, and target answer. We evaluate response correctness, timing, invalid outputs, interruption handling, and context continuity using Interaction-Aware Quality-Timeliness F1, Interruption Diagnostic Suite, and Nested Chain Completion Score. Experiments show that current models remain weak in streaming interaction, with the best overall IA-QTF1 reaching only 0.368 and the best 1QnA IA-QTF1 only 0.052. Further study on mathematical reasoning in full-duplex settings shows that offline capability does not necessarily transfer to online interaction. Code and datasets will be made publicly accessible at \url{https://github.com/Lucky-Lance/OmniInteract}.
\end{abstract}

%% file: src/1_intro.tex
\section{Introduction}\label{sec:intro}

Human--AI interaction is shifting from offline multimodal understanding to continuous, real-time communication~\citep{chen2025livecc,zeng2026streamforest,yang2025streamagent,liu2026thinking,xia2025streaming,fu2025vispeak,liu2024streamchat}. Conventional video-language evaluation typically asks models to answer questions after the relevant content has already been observed~\citep{fu2025video,li2024mvbench,wu2024longvideobench}, while recent streaming video benchmarks move closer to online perception~\citep{lin2026streamingbench,niu2025ovo,lu2026phostream}. Meanwhile, omnimodal large language models (LLMs) are integrating vision, audio, speech, and text into unified systems~\citep{chen2024far,chen2024internvl,team2026qwen3,comanici2025gemini,ai2025ming,cui2026minicpm}. These developments call for an evaluation setting beyond hindsight understanding: a real-time assistant must decide whether to respond, when to respond, and what to say during an ongoing audio-visual interaction.

\begin{figure*}[t]
    \centering
    \includegraphics[width=\linewidth]{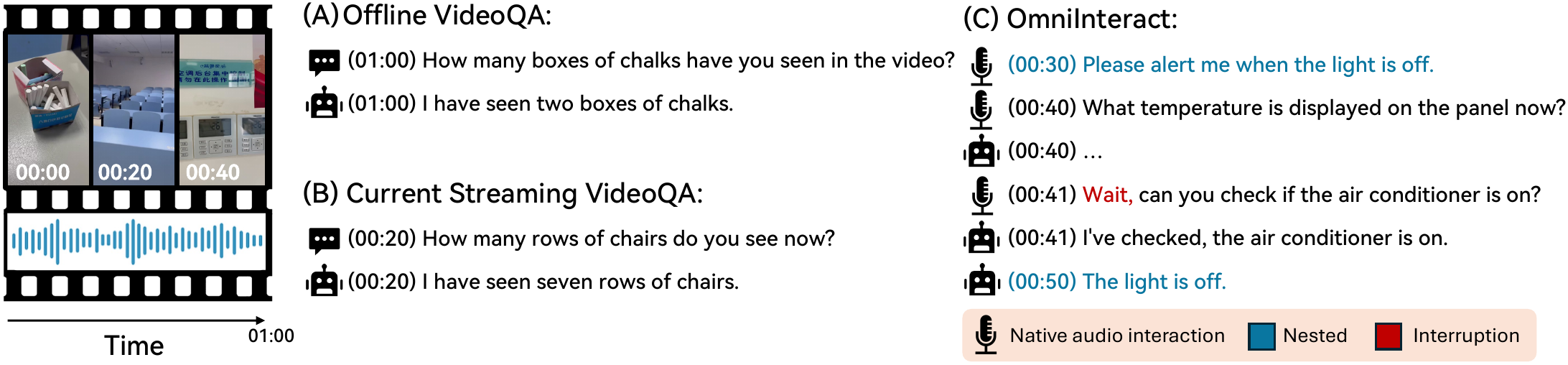}
    \vspace{-2em}
    \caption{Comparison of offline video QA, text-prompted streaming video QA, and OmniInteract (1Q1A). OmniInteract preserves spoken queries and multimodal events in the original audio-visual stream for timely, interruption-aware, and nested interaction evaluation.}
    \label{fig:comparison-qa}
    \vspace{-0.5em}
\end{figure*}

\begin{figure}
    \centering
    \includegraphics[width=\linewidth]{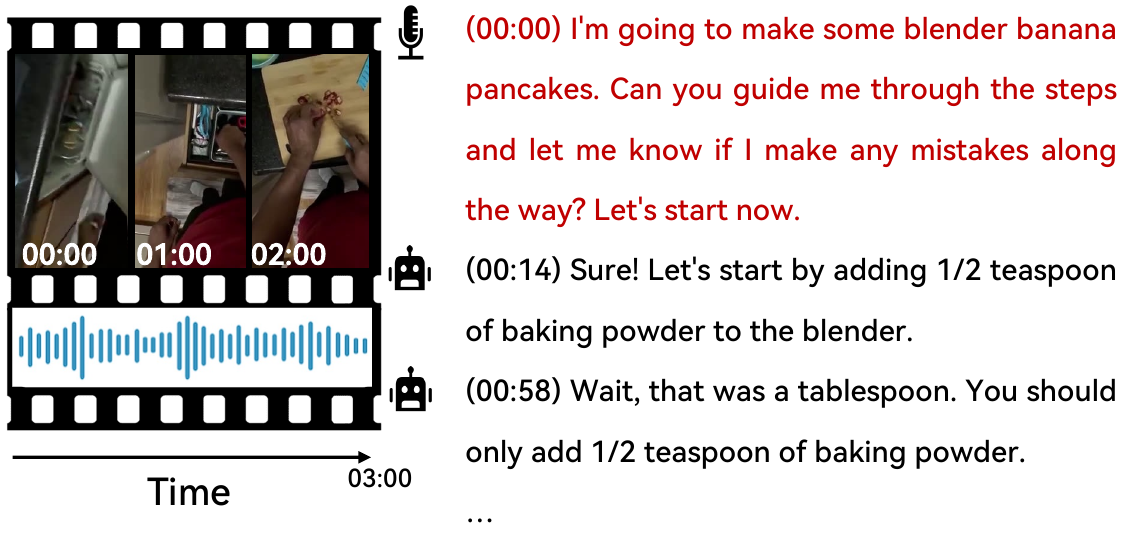}
    \vspace{-1.5em}
    \caption{Example of a 1QnA interaction. A single spoken instruction can require multiple temporally grounded response slots as the task unfolds.}
    \label{fig:1qna-example}
    \vspace{-0.5em}
\end{figure}

However, existing benchmarks do not fully capture this coupled decision process. Offline video question answering removes the need to decide response timing by allowing models to access the full video before answering~\citep{fu2025video,li2024mvbench,wu2024longvideobench,hu2025video,zhao2025mmvu}. Most streaming video benchmarks retain temporal inputs, but provide user questions as external textual prompts~\citep{lin2026streamingbench,niu2025ovo,lu2026phostream,wang2025omnimmi,wang2025proactivevideoqa}, bypassing spoken intent recognition from the audio stream. Moreover, existing benchmarks are evaluated on pre-segmented video clips with offline inference, or rely on custom streaming protocols distinct from the models' native real-time inference. As a result, they only partially evaluate the interaction loop required by native real-time assistants: detecting spoken or multimodal triggers, grounding them in visual events and background sounds, responding at the right moment, and avoiding invalid outputs while operating under genuine online streaming constraints. This limitation becomes more evident in full-duplex-oriented scenarios, where users may interrupt, insert new questions, or expect the assistant to resume an unfinished interaction~\citep{defossez2024moshi,yao2025flm,lin2025fullv1,lin2025fullv2,lin2026fullv3,cui2026minicpm}.

To evaluate this missing interaction loop, we introduce \textbf{OmniInteract}, a benchmark that directly evaluates omnimodal LLMs through their native online streaming inference in continuous real-time audio-visual streams. Fig.~\ref{fig:comparison-qa} contrasts this setting with offline and text-prompted streaming video QA. Rather than converting interactions into video-text question-answer pairs, OmniInteract preserves them in their native multimodal form: spoken user queries remain in the audio track, while visual events and background sounds remain part of the evolving context. Models must process the stream as it unfolds, without lookahead to future content. This design better reflects real interaction, but it also raises a practical question: how can a continuous audio-visual stream be evaluated when it does not naturally provide fixed question-answer boundaries?\looseness=-1

We address this question with an interaction slot formulation. Each slot represents a temporally grounded response opportunity, defined by a trigger, an expected response window, and a target answer. These elements correspond to the three key decisions in real-time interaction: the trigger indicates whether a response opportunity exists, the response window specifies when the model should answer, and the target answer defines what it should say. In this way, the slot formulation makes continuous omnimodal interaction measurable while preserving its temporal and multimodal nature.

Building on this formulation, OmniInteract includes two complementary interaction structures with 250 videos and 1,430 temporally grounded response slots in total. The \textbf{1Q1A} split contains 1,062 single-response slots (210 videos), including 638 real-time, 184 proactive, and 240 nested slots. It focuses on localized interactions constructed from self-recorded videos and manual annotations, where each trigger corresponds to one expected answer. The \textbf{1QnA} split contains 368 response slots  (40 videos) for continuous task monitoring from existing benchmarks, where a single instruction may require multiple temporally grounded responses as the task progresses; Fig.~\ref{fig:1qna-example} shows a representative example. Together, these splits evaluate whether models can handle both immediate response opportunities and longer-horizon monitoring within the original audio-visual stream.

The slot formulation also guides the evaluation metrics. Since each slot specifies both answer content and a valid response window, answer accuracy alone is insufficient: a semantically correct response may still fail as an interaction if it is produced too early, too late, or outside the intended context. OmniInteract further stresses interaction control with 192 interrupted response slots, including 147 in 1Q1A and 45 in 1QnA, as well as 240 nested slots forming 120 pairs that require models to answer an inserted inner query before resuming the outer query. We therefore propose an \textbf{Interaction-Aware Quality-Timeliness F1} (IA-QTF1), together with \textbf{Interruption Diagnostic Suite} (IDS) and the \textbf{Nested Chain Completion Score} (NCCS), to jointly measure response quality, timing, undesirable outputs, interruption handling, and context resumption.

\input{tabs/dataset_cmp}

We evaluate representative omnimodal real-time interaction models on OmniInteract. The results reveal substantial variation across scenarios, with continuous task monitoring remaining the most challenging setting because models must produce multiple temporally grounded responses over an extended stream. We further conduct a focused offline-online comparison on MiniCPM-o 4.5 mathematical reasoning tasks in a full-duplex-oriented setting~\citep{cui2026minicpm}, showing that reasoning quality degrades substantially when the model must reason while simultaneously listening and generating responses. Together, these results highlight a key gap in current omnimodal real-time interaction: strong multimodal understanding or reasoning in offline settings does not necessarily translate into robust real-time interaction.

Our contributions are summarized as follows:

\textbf{1)} We introduce \textbf{OmniInteract}, a benchmark for evaluating omnimodal LLMs through their native online streaming inference over continuous real-time audio-visual streams. OmniInteract preserves spoken queries, visual events, and background sounds in the original stream, and covers two complementary interaction structures: \textbf{1Q1A} for localized single-response interactions and \textbf{1QnA} for continuous task monitoring.

\textbf{2)} We propose an interaction slot formulation that represents each temporally grounded response opportunity with a trigger, an expected response window, and a target answer. Built on this, we develop Interaction-Aware Quality-Timeliness F1, Interruption Diagnostic Suite, and Nested Chain Completion Score, enabling joint evaluation of response content, timing, undesirable outputs, interruption handling, and context resumption.

\textbf{3)} We conduct a systematic benchmark analysis of representative omnimodal real-time interaction models under native spoken-query, online audio-visual interaction, with additional analyses of full-duplex-oriented behaviors. Our results reveal substantial gaps in current models, especially in continuous task monitoring and temporally grounded interaction control.

%% file: tabs/dataset_cmp.tex
\begin{table*}[t]\small
\centering
\renewcommand{\arraystretch}{0.95}
\caption{Benchmark comparison. We compare input modalities, query form, online inference, and interaction coverage across prior streaming video benchmarks and OmniInteract.}
\label{tab:benchmark_comparison}
\vspace{-0.5em}
\resizebox{\textwidth}{!}{
\begin{tabular}{l c c c c c c c c}
\toprule
\textbf{Benchmark} 
& \textbf{Modality} 
& \textbf{Query Modality} 
& \textbf{Online}
& \multicolumn{3}{c}{\textbf{1Q1A}} 
& \textbf{1QnA} 
& \textbf{Interruption} \\
\cmidrule(lr){5-7}
& & 
& \textbf{Inference}
& \textbf{Real-time} 
& \textbf{Proactive} 
& \textbf{Nested} 
& & \\
\midrule
StreamingBench~\cite{lin2026streamingbench} 
& V, A 
& T 
& \xmark 
& \cmark 
& \cmark 
& \xmark 
& \xmark 
& \xmark \\

OVO-Bench~\cite{niu2025ovo} 
& V 
& T 
& \xmark 
& \cmark 
& \cmark 
& \xmark 
& \xmark 
& \xmark \\

OmniMMI~\cite{wang2025omnimmi} 
& V, A 
& T 
& \xmark 
& \cmark 
& \cmark 
& \xmark 
& \xmark 
& \xmark \\

ProactiveVideoQA~\cite{wang2025proactivevideoqa} 
& V, A 
& T 
& \xmark 
& \xmark 
& \cmark 
& \xmark 
& \xmark 
& \xmark \\

PhoStream~\cite{lu2026phostream} 
& V, A 
& T 
& \cmark\textsuperscript{*}
& \cmark 
& \cmark 
& \xmark 
& \xmark 
& \xmark \\

\midrule
\textbf{OmniInteract (Ours)} 
& \textbf{V, A} 
& \textbf{A} 
& \textbf{\cmark} 
& \textbf{\cmark} 
& \textbf{\cmark} 
& \textbf{\cmark} 
& \textbf{\cmark} 
& \textbf{\cmark} \\
\bottomrule
\end{tabular}}
\vspace{0.2em}
\\
\raggedright \footnotesize 
\textbf{V}: Video, \textbf{A}: Audio, \textbf{T}: Text. 
\textsuperscript{*}: uses a custom streaming evaluation protocol rather than models' native online streaming inference.\looseness=-1
\end{table*}

%% file: src/2_related.tex
\section{Related Work}\label{sec:related}

\subsection{Streaming Video Understanding}

Streaming video understanding shifts from offline post-hoc understanding~\citep{fu2025video,li2024mvbench,wu2024longvideobench} to real-time online interaction~\citep{lin2026streamingbench,niu2025ovo,lu2026phostream,shen2026simple}, requiring synchronized perception, decision-making, and response. Recent works address this challenge through temporally aligned long-context modeling~\citep{chen2024videollm}, streaming token management with compact visual-text windows~\citep{xu2025streamingvlm}, asynchronous perception-decision-reaction pipelines~\citep{qian2025dispider}, proactive response training with dynamic compression~\citep{zhangeyes}, multi-turn reinforcement learning for timely responses~\citep{wang2025mmduet2}, offline-to-streaming adaptation with memory and activation mechanisms~\citep{wang2026streambridge}, and end-to-end continuous observation frameworks~\citep{lu2026aura}. These systems make important progress toward online video understanding, but existing benchmarks still only partially capture native real-time interaction. As summarized in Tab.~\ref{tab:benchmark_comparison}, they typically provide user queries as text rather than spoken audio, and evaluate models on pre-segmented clips using offline inference or custom streaming protocols instead of the models' native online streaming inference. These choices decouple response generation from the real-time perception, spoken intent recognition, and timing control required by native streaming assistants.

\subsection{Omnimodal Large Language Models}

Beyond temporal streaming, omnimodal LLMs extend multimodal interaction by integrating vision, audio, speech, and text within unified systems. Recent models add audio encoders to visual-language backbones~\citep{chen2024far,chen2024internvl}, unify multiple modalities in shared token spaces~\citep{team2026longcat}, scale native audio-visual interaction with mixture-of-experts and speech-generation architectures~\citep{team2026qwen3,ai2025ming}, and advance long-context multimodal reasoning over audio-visual inputs~\citep{comanici2025gemini}. These developments enable richer interaction interfaces, where user intent may appear as speech, background sounds may affect the response context, and visual events may determine when the model should answer. However, evaluation has not fully kept pace with these capabilities. Prior benchmarks cover parts of streaming video understanding, such as real-time or proactive QA, but they generally retain text queries, omit nested or multi-answer interaction structures, and do not evaluate interruption handling under native online inference. OmniInteract targets this gap by combining spoken audio queries, online model execution, 1Q1A and 1QnA interaction structures, and interruption-aware evaluation within the same benchmark.

\subsection{Full-Duplex Real-Time Interaction}

Streaming video understanding and omnimodal modeling naturally motivate full-duplex real-time interaction, where models process incoming input while generating output for more natural human--AI communication. Early full-duplex studies focus mainly on spoken dialogue, enabling low-latency speech-to-speech interaction without explicit turn segmentation~\citep{defossez2024moshi} and improving native audio interaction through dedicated training paradigms~\citep{yao2025flm}. Full-Duplex-Bench evaluates capabilities such as interruption handling, smooth turn-taking, and conversational continuity~\citep{lin2025fullv1,lin2025fullv2,lin2026fullv3}. At the multimodal level, recent work introduces a time-aligned streaming framework for simultaneous perception, speech generation, and proactive behavior~\citep{cui2026minicpm}. These works highlight the importance of interruption handling, overlapping input/output, and context continuation. OmniInteract complements them by evaluating such behaviors in continuous audio-visual streams with temporally grounded spoken-query interactions.

%% file: src/3_method.tex
\section{OmniInteract Benchmark}\label{sec:method}

\subsection{Data Composition}\label{sec:data_comp}

OmniInteract is constructed to evaluate omnimodal LLMs through their native online streaming inference in continuous real-time interaction scenarios. Unlike conventional offline video question answering~\citep{fu2025video,fu2026video}, where responses are produced after observing a complete video or clip, OmniInteract requires models to process the audio-visual stream as it unfolds, without lookahead to future content. We organize the data around interaction slots, each associated with a trigger, an expected response window, and a target answer (detailed in Sec.~\ref{sec:slot}). Beyond temporal streaming, OmniInteract further differs from prior streaming video benchmarks that often provide user questions as external textual inputs~\citep{lin2026streamingbench,niu2025ovo,lu2026phostream}. OmniInteract preserves the original audio-visual stream as the primary interaction context, where user queries are directly recorded in the audio track together with background sounds and visual events. This formulation evaluates whether models can recognize spoken intents, interpret multimodal evidence, and respond at appropriate moments in an end-to-end omnimodal setting.

Following this formulation, we categorize interaction instances according to whether they require a single response or multiple temporally evolving responses. OmniInteract is therefore organized into two complementary splits: 1Q1A and 1QnA. The 1Q1A split consists of instances where each trigger corresponds to one expected answer, and is further divided into three interaction types. Real-time interaction involves an explicit user query issued during the multimodal stream, where the model is expected to respond immediately based on the available context. Proactive interaction is driven by salient multimodal events rather than an explicit query, requiring the model to continuously monitor the stream and respond only when sufficient evidence or a relevant cue emerges. Nested interaction occurs when a real-time query is inserted within the response window of a proactive interaction, requiring the model to address the inserted query while maintaining the context of the original interaction. The 1QnA split covers cases where a single query or instruction corresponds to multiple valid answers over time. It evaluates whether a model can provide temporally appropriate responses as new evidence appears in the stream, rather than reducing the interaction to one static answer.

\input{tabs/data_stat}

Tab.~\ref{tab:data_statistics} summarizes the resulting split sizes. The 1Q1A split contains 1,062 response slots across real-time, proactive, and nested interactions, while 1QnA contains 368 response slots. The 147 interruptions in 1Q1A and 45 interruptions in 1QnA are annotated as cross-cutting cases within these splits rather than as a separate interaction type.

\subsection{Data Curation}

Given the different interaction structures of 1Q1A and 1QnA, we adopt different curation strategies for the two splits. Due to the lack of datasets specifically designed for native real-time omnimodal interaction, we curate the 1Q1A split from scratch. We self-record 210 videos in two groups of scenarios. The first group covers daily-life interactions in Chinese, including home activities, gym exercises, museums, shopping, and other common situated interactions (150 videos). The second group covers English mathematical problem-solving, where the user asks questions while the visual stream shows the evolving problem context (60 videos). For real-time interactions, we record explicit spoken queries in the audio track and align each query with the visual evidence needed for answering. For proactive interactions, the user first issues a spoken query whose answer is not yet available; the model must monitor the subsequent audio-visual stream and respond once the required evidence emerges. For nested interactions, we insert a real-time query into the response window of an ongoing proactive interaction, so that the model must answer the inserted query before resuming the original context. For each slot, we manually annotate the trigger, valid response window, and target answer, and verify that the answer is supported by the corresponding audio-visual evidence.

For the 1QnA split, we construct continuous monitoring instances from existing procedural and task-oriented video benchmarks (40 videos), including live step-by-step task guidance~\citep{bhattacharyya2026can,peddi2024captaincook4d} and egocentric error detection~\citep{lee2024error}. These sources naturally contain long-horizon activities in which multiple response opportunities arise as the task progresses. Starting from the original task goal, step annotations, and temporal event labels, we convert each example into an interaction stream with one initial instruction and multiple response slots. Specifically, we rewrite the task topic or goal into a natural user instruction, synthesize it into speech using text-to-speech~\citep{hu2026qwen3}, and prepend the synthesized instruction to the original audio-visual stream. We then map step-level guidance targets or error events to temporally grounded response slots, each with its own answer time and target response. This procedure preserves the original video evidence while turning offline task annotations into an end-to-end audio-visual interaction setting, where the model receives the instruction through audio and must decide when to respond as new evidence appears. Benchmark examples are shown in Fig.~\ref{fig:comparison-qa} (1Q1A) and Fig.~\ref{fig:1qna-example} (1QnA).

\subsection{Evaluation Metrics}
Continuous real-time human--AI interaction shifts evaluation from static correctness to dynamic interaction management. Traditional metrics are insufficient for online settings, particularly for handling full-duplex interruptions and nested context resumption. We therefore build our scoring framework upon the interaction slot formulation, anchoring evaluation to the triggers, response windows, and target answers introduced in Sec.~\ref{sec:slot} to jointly measure response timeliness, content quality, and conversational continuity.

\begin{figure*}[t]
    \centering
    \includegraphics[width=0.95\linewidth]{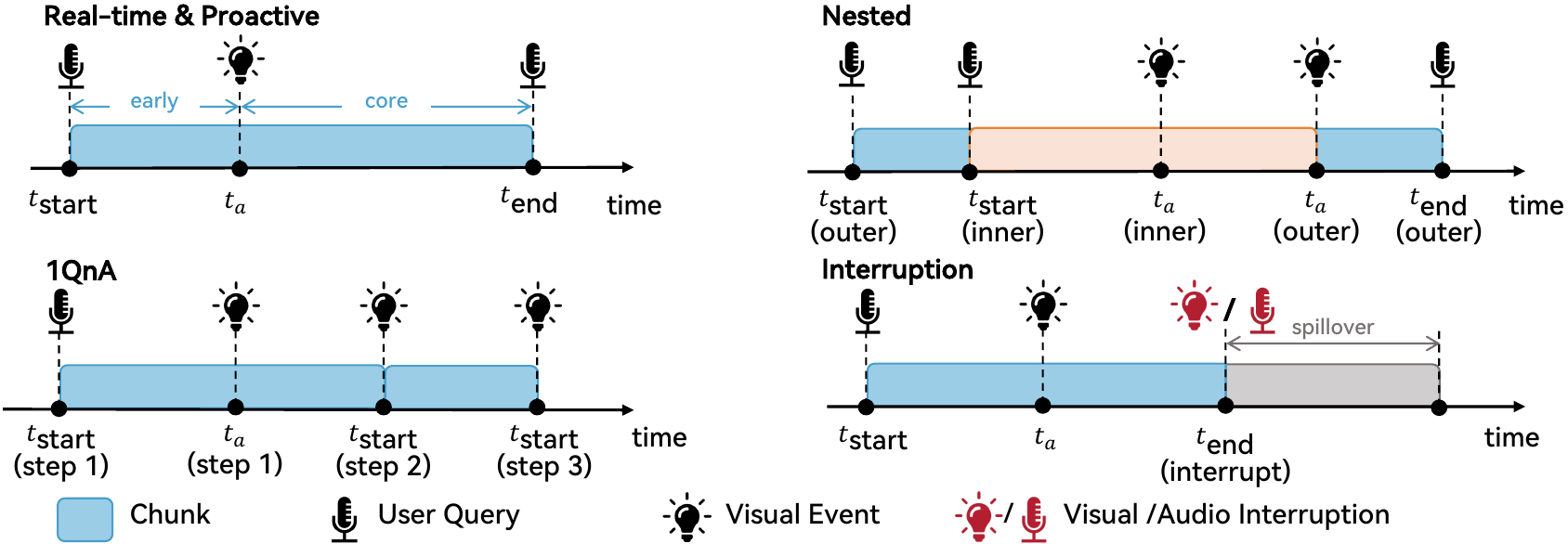}
    \caption{Interaction slot construction for real-time, proactive, nested, 1QnA, and interruption settings. Generated chunks are assigned to temporal slots and split into early and core segments around the valid-answer time for interaction-aware evaluation.}
    \label{fig:slot}
    \vspace{-1em}
\end{figure*}

\subsubsection{Slot Construction and Chunk Matching}\label{sec:slot}
Continuous streams do not provide explicit turn boundaries, so we discretize evaluation into interaction slots:
\begin{equation}
    \text{slot} =[t_{\text{start}}, t_a, t_{\text{end}}),
\end{equation}
where $t_{\text{start}}$ is the onset of observation, $t_a$ is the earliest moment for a valid core response, and $t_{\text{end}}$ is the window's close. 
Fig.~\ref{fig:slot} illustrates how slots are constructed across representative interaction types defined in Sec.~\ref{sec:data_comp}.

We establish real-time and proactive interactions as the foundational structure: $t_{\text{start}}$ aligns with the user query, $t_a$ is the time of the visual event that enables a valid answer, and $t_{\text{end}}$ is bounded by the subsequent query. For nested interactions, the outer slot keeps this definition, while the inserted query opens an inner slot that ends at $t_a(\text{outer})$, when the visual event makes the outer proactive response timely again and evaluation switches back to the outer slot. 
For 1QnA, which handles sequential responses to a single instruction, the first step follows the foundational structure. In subsequent steps, each visual event triggers the next slot, whose $t_{\text{start}}$ and $t_a$ align (labeled as $t_{\text{start}}$), and the next slot's $t_{\text{start}}$ serves as current slot's $t_{\text{end}}$.
Within these settings, a new user query or visual event of another slot (which defines $t_{\text{end}}$) may arrive before the current answer is completed. We refer to this as an interruption, where the current slot is termed the interrupted slot, completing its response is not required, and any output after $t_{\text{end}}$ is considered spillover. In practice, we annotate an interruption when the interval $[t_a,t_{\text{end}})$ is shorter than the TTS-estimated duration of the ground-truth answer.

Building on these definitions, a model-generated text chunk is assigned to a slot if its start time falls within $[t_{\text{start}},t_{\text{end}})$. In cases of overlap, such as nested resumptions, the chunk is mapped to the slot with the latest $t_{\text{start}}$ time, prioritizing the most recent context. Chunks straddling the $t_a$ boundary are split at the word level into an early segment (before $t_a$) and a core segment (from $t_a$ onward). Unassigned chunks are recorded as unmatched outputs and penalized during metric computation.

\subsubsection{Interaction-Aware Scoring}
For each slot, we compute a unified set of stage-specific scores to derive soft true positives ($TP$) and discrete penalties ($FP$, $FN$), integrating interaction management into a generalized framework.

\textbf{Stage-Specific Scoring.} We evaluate intra-slot outputs across an early stage ($t < t_a$) and a core stage ($t \ge t_a$), both incorporating a time-decay mechanism to reward promptness. \textbf{1)} The early stage evaluates tentative acknowledgments or feedback, where valid interactions are rewarded based on onset timing, while early hallucinations yield zero. \textbf{2)} The core stage assesses the correctness and coverage of the ground-truth answer, penalized by its latency relative to $t_a$. The total validity of an interaction is a soft true positive (TP), computed as the clamped sum of both stage scores. Full scoring definitions are provided in Appendix~\ref{sec:scoring_detail}.

\input{tabs/exp_f1}

\textbf{Classification and Global Metric.} Interaction failures are captured via discrete penalties. A false negative (FN) is assigned when a non-interruption slot lacks a core answer. A false positive (FP) aggregates four unwarranted behaviors: \textbf{1)} unmatched chunks, \textbf{2)} early hallucinations, \textbf{3)} low-quality responses, and \textbf{4)} spill, where output exceeds the boundary $t_{\text{end}}$ to disrupt conversational continuity. Across all slots, we define the Interaction-Aware Quality-Timeliness F1 (IA-QTF1) as:
\begin{equation}
    \text{IA-QTF1} = \frac{2 \cdot \sum TP}{2 \cdot \sum TP + \sum FP + \sum FN}.
\end{equation}
By using soft TP values to account for response timing while penalizing flow-breaking behaviors like spill, IA-QTF1 provides a comprehensive assessment of a model's ability to manage dynamic multimodal dialogue.

\subsubsection{Extended Metrics}
\label{sec:extended_metrics}
To further assess specific interaction capabilities in greater detail, we define targeted metrics for interruption handling and nested context management.

\textbf{Interruption Diagnostic Suite (IDS).} Interrupted slots include both user-initiated interruptions, where the original answer is often no longer needed, and event-triggered shifts, where partial answers to the preempted query may still be useful. Because Global IA-QTF1 treats all interruptions as boundary-control cases and does not reward incomplete answers to the preempted query, the metric does not distinguish between silence, useful partial responses, and post-interruption spillover. IDS addresses this gap with three complementary diagnostics: \textbf{No-Output Rate (NOR)}, the proportion of interrupted slots with no model output for the preempted query; \textbf{Partial Answer Quality (PAQ)}, an LLM-judged usefulness score for already-spoken content without incompleteness penalties; and \textbf{Conditional Spill Metrics (CSM)}, spill rate and average spill duration computed only over interrupted slots with output.

\textbf{Nested Chain Completion Score.} To evaluate state management during inserted queries, we further define the Nested Chain Completion Score (NCCS) as the geometric mean of correctness across the outer--inner query pair:
\begin{equation}
    \text{NCCS} = \sqrt{\text{Score}_{\text{outer}} \times \text{Score}_{\text{inner}}}\,.
\end{equation}
Here, $\text{Score}_{\text{outer}}$ and $\text{Score}_{\text{inner}}$ are outer/inner core-stage scores. NCCS requires answering the inner query and then resuming the outer query, measuring context-switching and resumption fidelity.
\looseness=-1

%% file: tabs/data_stat.tex
\begin{table}[t]
\centering
\small
\setlength{\tabcolsep}{4pt}
\caption{Statistics of OmniInteract. Video counts denote the number of source videos; slot counts denote temporally grounded response slots; interruptions are cross-cutting cases included in the corresponding split.}
\vspace{-0.5em}
\label{tab:data_statistics}
\begin{tabular}{lcccc}
\toprule
\textbf{Split} & \textbf{Type} & \textbf{Videos} & \textbf{Slots} & \textbf{Interruptions} \\
\midrule
\multirow{3}{*}{1Q1A} 
& Real-time & \multirow{3}{*}{210} & 638 & \multirow{3}{*}{147} \\
& Proactive &  & 184 &  \\
& Nested    &  & 240 &  \\
\midrule
1QnA & Monitoring & 40 & 368 & 45 \\
\midrule
\multicolumn{2}{l}{\textbf{Total}} & \textbf{250} & \textbf{1,430} & \textbf{192} \\
\bottomrule
\end{tabular}
\vspace{-1em}
\end{table}

%% file: tabs/exp_f1.tex
\begin{table*}[t]\small
\centering
\renewcommand{\arraystretch}{0.9}
\caption{IA-QTF1 across interaction settings. The 1Q1A columns use mutually exclusive real-time, proactive, and nested response slots; global scores are recomputed from aggregated TP/FP/FN.}
\vspace{-0.5em}
\label{tab:unified_final}

\begin{adjustbox}{max width=\textwidth}
\begin{tabular}{l cccc cc}
\toprule
\multirow{2}{*}{\textbf{Model}} & \multicolumn{4}{c}{\textbf{1Q1A}} & \multirow{2}{*}{\textbf{1QnA}} & \multirow{2}{*}{\textbf{All Global}} \\
\cmidrule(lr){2-5}
 & \textbf{Real-time} & \textbf{Proactive} & \textbf{Nested} & \textbf{1Q1A Global} & & \\
\midrule
AURA & 0.376 & 0.549 & 0.596 & \textbf{0.467} & \textbf{0.052} & 0.363 \\
Gemini 2.5 Flash Live & \textbf{0.553} & 0.121 & 0.398 & 0.428 & 0.028 & 0.344 \\
MiniCPM-o 4.5 & 0.337 & \textbf{0.607} & \textbf{0.599} & 0.456 & 0.015 & \textbf{0.368} \\
Qwen3.5-Omni Flash Realtime & 0.524 & 0.108 & 0.379 & 0.401 & 0.023 & 0.323 \\
\bottomrule
\end{tabular}
\end{adjustbox}
\vspace{-1em}
\end{table*}

%% file: src/4_exp.tex
\section{Experiments}\label{sec:exp}

We evaluate four representative omnimodal real-time models: AURA~\citep{lu2026aura}, Gemini 2.5 Flash Live~\citep{comanici2025gemini}, MiniCPM-o 4.5~\citep{cui2026minicpm}, and Qwen3.5-Omni Flash Realtime~\citep{team2026qwen3}. All models are tested using their original real-time inference pipelines and native audio-visual streams, requiring them to jointly handle spoken user intents, visual evidence, and response timing. Since the answers are open-ended, we use GPT-4o~\citep{hurst2024gpt} as an external judge to compare model responses against ground-truth annotations, thereby reducing evaluator bias from the tested models. The judge protocol is detailed in Appendix~\ref{sec:judge_protocol}.

\subsection{Inference Protocol}

Although OmniInteract is distributed as offline audio-visual recordings for reproducible evaluation, all models are evaluated under an online streaming protocol. During inference, each recording is replayed chronologically to the model through its native real-time interface, so that frames and audio are exposed only according to their original timestamps. The model can therefore condition on past and current inputs, but cannot access future video frames, future audio, or ground-truth slot boundaries. We timestamp model outputs during replay and align the generated chunks with interaction slots after inference using the procedure in Sec.~\ref{sec:slot}. This protocol simulates real online interaction while keeping the benchmark deterministic and comparable across models.

\subsection{1Q1A Interaction}

\input{tabs/exp_nested}
\input{tabs/exp_interruption}

\input{tabs/exp_degradation}

The 1Q1A split evaluates localized response opportunities, including explicit user queries, proactive triggers, and nested queries. Tab.~\ref{tab:unified_final} reports IA-QTF1 for each category and the global score. 

For explicit real-time queries, Gemini obtains the best score ($0.553$), followed by Qwen3.5-Omni ($0.524$), showing stronger performance when the user intent is directly stated. In contrast, proactive interaction favors MiniCPM-o ($0.607$) and AURA ($0.549$), suggesting better monitoring after an earlier query whose answer becomes available only later. On nested slots, MiniCPM-o and AURA again perform best, indicating stronger local handling of context shifts. Under the global 1Q1A metric, which aggregates TP/FP/FN across all slots, AURA achieves the highest IA-QTF1 ($0.467$), slightly ahead of MiniCPM-o ($0.456$).\looseness=-1

Nested IA-QTF1 measures local validity of inner and outer answers, but does not fully capture whether the model resumes the suspended outer query after the inserted query. We therefore report NCCS in Tab.~\ref{tab:nested}. MiniCPM-o achieves the best NCCS of 0.284, followed by AURA at 0.270. Although Gemini and Qwen3.5-Omni answer many inner queries correctly, they fail to resume the outer query in 119 and 116 of 120 cases, respectively, indicating that current models often treat nested queries as permanent context switches rather than temporary interruptions requiring resumption.

\subsection{1QnA Interaction}

The 1QnA split evaluates continuous task monitoring, where a single instruction may require multiple temporally grounded responses. As shown in Tab.~\ref{tab:unified_final}, all models perform substantially worse on 1QnA than on 1Q1A. AURA obtains the highest IA-QTF1 score of 0.052, but the absolute score remains low. This suggests that long-horizon interaction remains difficult, as models often miss intermediate response opportunities or respond at inappropriate times, even when they can handle isolated 1Q1A cases.

When aggregating both splits, MiniCPM-o obtains the highest overall Global IA-QTF1 score of $0.368$, followed by AURA at $0.363$. The small gap between the best models, together with the uniformly low 1QnA scores, suggests that current systems have not yet achieved robust general-purpose streaming interaction behavior across localized and long-horizon settings (detailed breakdown in Appendix~\ref{sec:breakdown}).\looseness=-1

\subsection{More Interruption Analyses}

We use the Interruption Diagnostic Suite (IDS) defined in Sec.~\ref{sec:extended_metrics} to further separate no output for the preempted query from failed stopping behavior and to measure conditional spill severity. Tab.~\ref{tab:interruption} shows that Gemini avoids spillover mostly through conservative silence, with the highest NOR (85.94\%), modest PAQ (0.370), and the best CSM (40.74\%, 0.312 s). MiniCPM-o shows the opposite pattern: it responds more often, with a lower NOR of 53.65\% and the best PAQ of 0.571, but spills severely when it responds, with CSM of 83.15\% and 10.067 s. Qwen3.5-Omni is more balanced, with NOR of 71.35\% and relatively low CSM of 41.82\% and 0.613 s, while AURA combines high silence (NOR 79.17\%) with modest PAQ (0.293) and elevated spillover (CSM 60.00\%, 1.879 s).

\subsection{Full-duplex Capability Degradation}

Finally, we examine whether offline capability transfers to online full-duplex-oriented interaction. We focus on MiniCPM-o 4.5, which is, to the best of our knowledge, the only open-source model that currently supports full-duplex real-time interaction. For offline inference, the entire question video is provided to MiniCPM-o at once, and the model answers after observing the full input. We compare its mathematical reasoning performance under offline inference and online full-duplex streaming interaction. To isolate answer correctness, we report the pure quality score (by GPT-4o), which excludes time decay and FP/FN penalties. As shown in Tab.~\ref{tab:degradation}, MiniCPM-o drops from $0.6833$ offline to $0.3475$ online, an absolute decrease of $0.3358$. This suggests that continuous listening, visual processing, and concurrent response generation can substantially degrade reasoning quality. This result reinforces the need to evaluate omnimodal models in native streaming interaction, rather than relying solely on offline multimodal reasoning scores, highlighting the value of OmniInteract as a benchmark.

%% file: tabs/exp_nested.tex
\begin{table*}[t]
\centering
\small
\setlength{\tabcolsep}{4pt}
\renewcommand{\arraystretch}{0.9}

\caption{Nested interaction results over 120 nested pairs. NCCS measures chain-level completion, while Inner and Outer IA-QTF1 report local slot quality.}
\label{tab:nested}
\vspace{-0.5em}
\begin{tabular*}{\textwidth}{@{\extracolsep{\fill}}lcccc@{}}
\toprule
\textbf{Model} 
& \textbf{NCCS} 
& \textbf{Inner IA-QTF1} 
& \textbf{Outer IA-QTF1} 
& \textbf{Missed Outer} \\
\midrule
AURA     
& 0.270 
& 0.595 
& 0.599 
& 54 / 120 \\

Gemini 2.5 Flash Live
& 0.001 
& 0.595 
& 0.165 
& 119 / 120 \\

MiniCPM-o 4.5 
& \textbf{0.284} 
& 0.587 
& \textbf{0.612} 
& 55 / 120 \\

Qwen3.5-Omni Flash Realtime
& 0.012 
& \textbf{0.702} 
& 0.092 
& 116 / 120 \\
\bottomrule
\end{tabular*}
\vspace{-1.5em}
\end{table*}

%% file: tabs/exp_interruption.tex
\begin{table}[t]
\centering
\footnotesize
\setlength{\tabcolsep}{3pt}
\renewcommand{\arraystretch}{0.9}
\caption{Interruption Diagnostic. NOR: No-Output Rate; PAQ: Partial Answer Quality; CSM-SR: Conditional Spill Rate; CSM-AS: Conditional Average Spill.}
\vspace{-2mm}
\label{tab:interruption}

\begin{tabular*}{\linewidth}{@{\extracolsep{\fill}}lcccc@{}}
\toprule
\textbf{Model}
& \textbf{NOR}
& \textbf{PAQ $\uparrow$}
& \textbf{CSM-SR $\downarrow$}
& \textbf{CSM-AS (s) $\downarrow$} \\
\midrule
AURA & 79.17\% & 0.293 & 60.00\% & 1.879 \\
Gemini & 85.94\% & 0.370 & \textbf{40.74\%} & \textbf{0.312} \\
MiniCPM-o & 53.65\% & \textbf{0.571} & 83.15\% & 10.067 \\
Qwen-Omni & 71.35\% & 0.361 & 41.82\% & 0.613 \\
\bottomrule
\end{tabular*}
\vspace{-0.5em}
\end{table}

%% file: tabs/exp_degradation.tex
\begin{table}[t]\small
\centering
\caption{Full-duplex capability degradation. We compare the mathematical reasoning quality of MiniCPM-o 4.5 in offline and online (full-duplex) settings.}
\vspace{-0.5em}
\label{tab:degradation}
\resizebox{\linewidth}{!}{
\begin{tabular}{lccc}
\toprule
\textbf{Metric} & \textbf{Offline} & \textbf{Online} & $\Delta$ \textbf{Drop} \\
\midrule
Pure Quality Score & 0.6833 & 0.3475 & -0.3358 \\
\bottomrule
\end{tabular}
}
\vspace{-1.5em}
\end{table}

%% file: src/5_conclusion.tex
\section{Conclusion}\label{sec:conclusion}

We introduced OmniInteract, a benchmark for evaluating omnimodal LLMs in native online streaming audio-visual interaction. Unlike offline or pre-segmented QA benchmarks, OmniInteract preserves spoken queries, visual events, ambient sounds, and response timing, enabling joint evaluation of answer quality, timeliness, interruption handling, and context resumption. Experiments show that current models struggle with robust real-time interaction, especially in long-horizon 1QnA monitoring and nested query resumption. These results highlight the gap between offline multimodal understanding and reliable full-duplex-oriented interaction, providing a foundation for future research on more natural human--AI communication.

\section*{Limitations}

OmniInteract has several limitations that point to future work. First, we evaluate four representative models, but the landscape of omnimodal systems is evolving rapidly. Second, the online capability degradation analysis is limited to MiniCPM-o on mathematical reasoning tasks. Third, the 1QnA split uses TTS-synthesized speech for initial instructions, while 1Q1A queries are naturally recorded, which may introduce variation in speech recognition difficulty. Finally, the benchmark currently covers Chinese daily-life interactions and English mathematical reasoning, and broader language and domain coverage remains future work.

\section*{Ethical Considerations}

OmniInteract is a research benchmark for evaluating real-time omnimodal interaction capabilities. It does not collect or release unauthorized personal user data; all self-recorded videos were created by the authors with informed consent from individuals who appear in them, and the 1QnA split builds on publicly available datasets under their original licenses. While real-time omnimodal assistants may support accessibility, education, and hands-free guidance, always-on multimodal systems also raise privacy and surveillance concerns that require careful deployment safeguards.

%% file: src/x_appendix.tex
\clearpage
\appendix
\raggedbottom

\section{Appendix}\label{sec:appendix}

\setcounter{figure}{0}
\setcounter{table}{0}
\setcounter{lstlisting}{0}

\renewcommand{\thefigure}{\thesection.\arabic{figure}}
\renewcommand{\thetable}{\thesection.\arabic{table}}
\renewcommand{\thelstlisting}{\thesection.\arabic{lstlisting}}

\subsection{Data Licenses and Annotation Details}\label{sec:data_license}

Tab.~\ref{tab:data_license} summarizes the licenses and access terms for the external data sources and data-generation tools. For human annotation, annotators were compensated at a rate of US\$20 per hour.

\input{tabs/data_license}

\input{tabs/detail_tab}

\subsection{Detailed Scoring Definitions}\label{sec:scoring_detail}

\textbf{Early Stage Score ($\text{Score}_{\text{ack}}$).}
Within the early segment $[t_{\text{start}}, t_a)$, model outputs are evaluated for appropriate interaction behavior. Valid acknowledgments (\textit{e.g.}, confirmations, brief feedback, or wait signals) are rewarded with a score that decays with onset latency relative to the early window length, scaled by a cap factor $\alpha$ to ensure that acknowledgments contribute less than core answers. If the model instead produces an early hallucination (\textit{i.e.}, a substantive answer before sufficient evidence has emerged at $t_a$), the acknowledgment score is set to zero and a false positive is recorded.

\textbf{Core Stage Score ($\text{Score}_{\text{core}}$).}
Within the core segment $[t_a, t_{\text{end}})$, the score combines a semantic quality factor and a timeliness factor:
\begin{equation}
    \text{Score}_{\text{core}} = S_{\text{core}} \times T_{\text{core}},
\end{equation}
where $S_{\text{core}} \in [0,1]$ is the semantic quality score assigned by the LLM judge (Sec.~\ref{sec:judge_protocol}), assessing correctness and coverage against the ground-truth answer. $T_{\text{core}} \in [0,1]$ is a timeliness factor that decays linearly from $1$ to $0$ as the semantic anchor (\textit{i.e.}, the earliest chunk containing the key answer content, as identified by the judge) shifts from $t_a$ toward $t_{\text{end}}$:
\begin{equation}
    T_{\text{core}} = \max\!\left(0,\; 1 - \frac{t_{\text{anchor}} - t_a}{t_{\text{end}} - t_a}\right).
\end{equation}

\textbf{Soft True Positive.}
The per-slot soft TP is defined as:
\begin{equation}
    TP_n = \min\!\left(1,\;\text{Score}_{\text{ack}} + \text{Score}_{\text{core}}\right)\,.
\end{equation}
The clamping ensures the combined score does not exceed~$1$.

\textbf{False Positive Categories.}
Each slot may incur FP counts from four sources: (1)~unmatched chunks not assigned to any slot, (2)~early hallucinations in the $[t_{\text{start}}, t_a)$ segment, (3)~core responses with quality below a minimum threshold, and (4)~spillover output beyond the slot boundary $t_{\text{end}}$. \looseness=-1

\textbf{False Negative.}
A non-interrupted slot is assigned $FN\!=\!1$ when $\text{Score}_{\text{core}} \le 0$, \textit{i.e.}, the model fails to produce any valid core answer within the response window. Acknowledgments alone do not satisfy the completion requirement. Interrupted slots do not incur FN, since the interaction was preempted before the model was expected to complete its answer.

\textbf{More Interruption Diagnostics.}
For interrupted slots, the global IA-QTF1 score only checks boundary control and does not require completing the original answer. We therefore report separate diagnostics. \textbf{No-Output Rate (NOR)} is the fraction of interrupted slots with no model output. For interrupted slots with output, \textbf{Partial Answer Quality (PAQ)} is an LLM-judged score in $[0,1]$ measuring whether the already spoken partial response is relevant, correct, and useful; incompleteness alone is not penalized. \textbf{Conditional Spill Metrics (CSM)} measure spill rate and average spill duration only over interrupted slots with output. %

\subsection{Detailed TP/FP/FN Breakdown}\label{sec:breakdown}

Tab.~\ref{tab:breakdown} reports the per-category TP, FP, and FN values underlying the IA-QTF1 scores in Tab.~\ref{tab:unified_final}. The 1Q1A categories use mutually exclusive response slots; the global score aggregates all $1{,}430$ slots and includes unmatched-chunk FP that are not attributed to any individual category.

\clearpage
\onecolumn

\begin{multicols}{2}

\subsection{LLM Judge Evaluation Protocol}\label{sec:judge_protocol}

All open-ended answer assessments use GPT-4o~\citep{hurst2024gpt} as an external judge to avoid evaluator bias from the tested models. Core-stage assessment receives: (1)~the ground-truth target answer, (2)~the concatenated model-generated chunks within the core segment, and (3)~a structured instruction asking it to rate semantic correctness and coverage on a continuous scale of $[0, 1]$. The judge also identifies the semantic anchor (\textit{i.e.}, the earliest chunk that contains the key answer content) used to compute the timeliness factor $T_{\text{core}}$. Early-stage assessment is performed separately on chunks before $t_a$, where the judge classifies outputs as either valid acknowledgments (brief interaction feedback) or early hallucinations (premature substantive content). For 1QnA slots, the judge additionally checks whether the model reveals information about future steps before they become relevant, flagging such outputs as spoilers. For interrupted slots with output, the judge scores the Partial Answer Quality (PAQ) of the already spoken content without penalizing incompleteness. 

\columnbreak

The judge uses separate prompts for early-stage, core-stage, and interruption-diagnostic scoring. Listing~\ref{lst:early_prompt} classifies outputs in $[t_{\text{start}}, t_a)$ as neutral acknowledgments or early hallucinations. Listing~\ref{lst:interrupt_partial_prompt} scores the usefulness of already spoken partial outputs without penalizing incompleteness. Listing~\ref{lst:core_prompt} scores core-answer quality in $[t_a,t_{\text{end}})$ and extracts the trigger phrase for semantic anchor identification. The prompts shown below are English translations of the original Chinese prompts used in evaluation.

\end{multicols}

\begin{nolinenumbers}
\begin{lstlisting}[caption={Early-stage judge prompt template.},label={lst:early_prompt},frame=single]
[System]
You are a streaming voice assistant evaluation judge. Judge only based on the given text. Output must be parseable JSON with no other text.

[User]
Determine whether the early output between start and t_a is an early hallucination.

[scene_type] {scene_type}
[slot] slot_id={slot_id},
  turn_index={turn_index},
  step_index={step_index},
  boundary_type={boundary_type},
  is_interrupted={is_interrupted}
[question] {question}
[current_gt_answer] {gt_answer}
[full_chunk_context] {full_context}
[early_actual_text] {actual_text}

Rules:
1. Greetings, confirmations, waiting, brief observations, and follow-up phrases -> Neutral.
2. If the model starts substantively answering, guessing unseen info, revealing future steps, or making definitive factual claims -> FP.
3. For 1QnA first step, reciting the full procedure before acting -> FP.
4. score is interaction quality 0-1 when Neutral; 0 when hallucination.

Output JSON:
(*@\{@*)"flag":"Neutral|FP_Hallucination",
 "score":float 0-1,
 "rationale":"one sentence"(*@\}@*)
\end{lstlisting}

\begin{lstlisting}[caption={Interrupted partial-answer quality prompt template.},label={lst:interrupt_partial_prompt},frame=single]
[System]
You are a strict evaluator for interrupted voice-assistant answers. Judge only from the provided text. Return valid JSON only, with no extra text.

[User]
Evaluate the quality of the assistant output that was already spoken before or around an interruption.

[Task]
The assistant was answering, but the interaction was interrupted before completion. The assistant was not required to complete the full original answer. Score whether the content already spoken is relevant, correct, and useful for the current ground-truth answer.

[Question] {question}
[Ground Truth Answer] {gt_answer}
[Assistant Output Already Spoken]
{actual_text}

Scoring Rules:
1. Score from 0 to 1.
2. Do not penalize incompleteness: a partial answer can receive a high score if the spoken part is correct and useful.
3. Score high when the spoken content overlaps with, paraphrases, or conveys useful parts of the ground truth.
4. Score low for acknowledgments or prefaces without substantive answer content.
5. Score low for wrong-question, irrelevant, or generic-filler output.
6. hallucination=true if the output contains clear incorrect facts, wrong target content, or unsupported content.
7. Ignore overflow duration when scoring quality; spill is measured separately.

Output JSON:
(*@\{@*)"score":float 0-1,
 "hallucination":true|false,
 "rationale":"one sentence"(*@\}@*)
\end{lstlisting}

\begin{lstlisting}[caption={Core-stage judge prompt template.},label={lst:core_prompt},frame=single]
[System]
You are a strict streaming voice assistant core-answer evaluation judge. Judge only based on the given text and reference answer. Output must be parseable JSON with no other text.

[User]
Score the core output after t_a.

[scene_type] {scene_type}
[slot] slot_id={slot_id},
  turn_index={turn_index},
  step_index={step_index},
  boundary_type={boundary_type},
  is_interrupted={is_interrupted}
[question] {question}
[current_gt_answer] {gt_answer}
[future_gt_answers_or_steps]
  {future_answers}
[full_chunk_context] {full_context}
[core_actual_text] {actual_text}

Rules:
1. score 0-1: correctness and coverage of core_actual_text vs gt_answer.
2. Off-topic, factual errors, or missing key answer -> low score.
3. 1QnA: reward only current-step info; penalize spoiling future steps or skipping the current step.
4. If score > 0, extract the earliest contiguous substring from core_actual_text that establishes the answer as trigger_phrase.
5. trigger_phrase must be a verbatim substring; empty if score == 0.

Output JSON:
(*@\{@*)"score":float 0-1,
 "trigger_phrase":"substring or empty",
 "spoiler":true|false,
 "rationale":"one sentence"(*@\}@*)
\end{lstlisting}

\end{nolinenumbers}
\twocolumn

\clearpage

\subsection{Case Study}\label{sec:case_study}

\begin{figure*}[t]
    \centering
    \includegraphics[width=\linewidth,height=0.95\textheight,keepaspectratio]{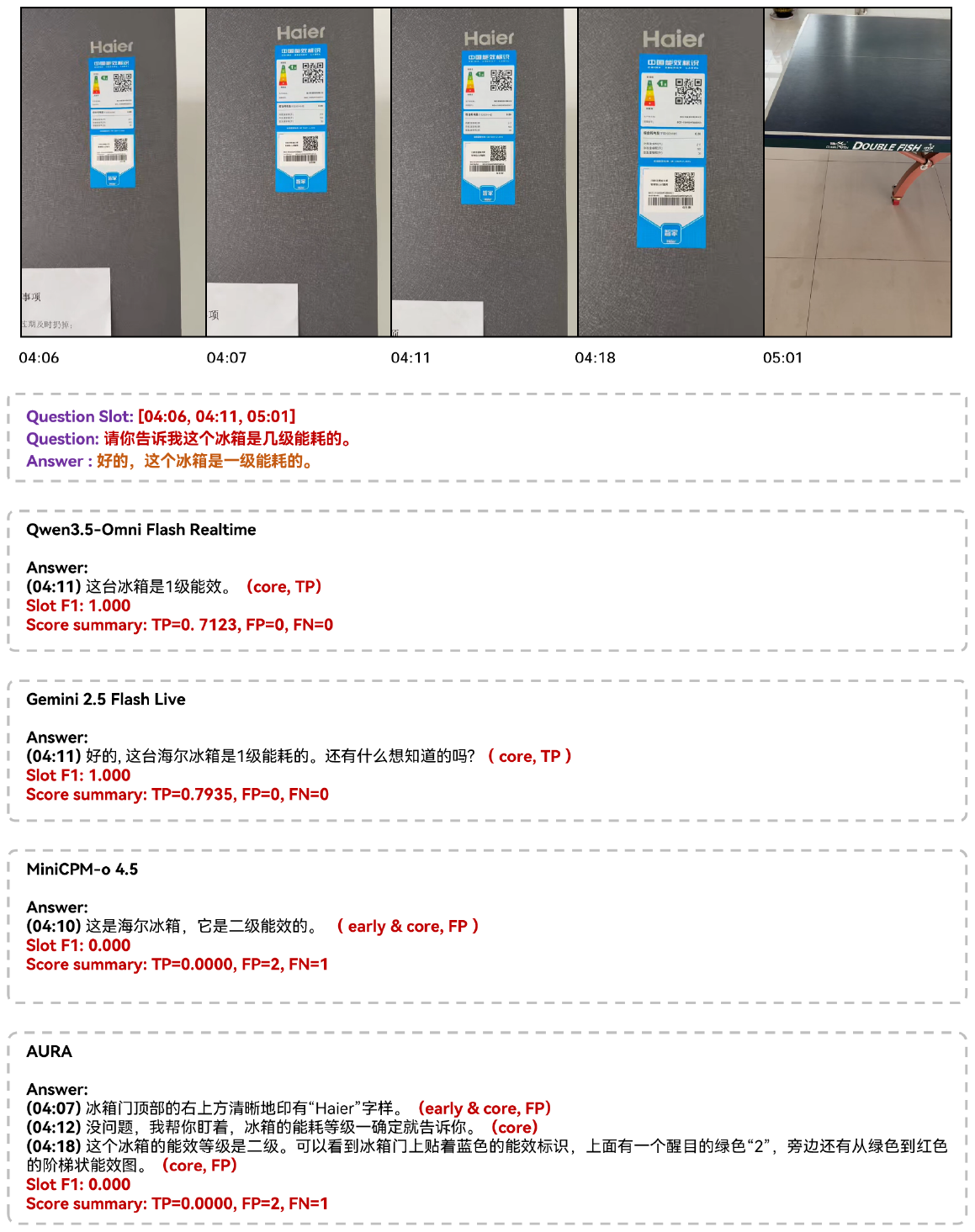}
    \caption{Real-time interaction case. The user asks for the energy-efficiency level of a Haier refrigerator after the label becomes visible.}
    \label{fig:case_realtime}
\end{figure*}

\begin{figure*}[t]
    \centering
    \includegraphics[width=\linewidth,keepaspectratio]{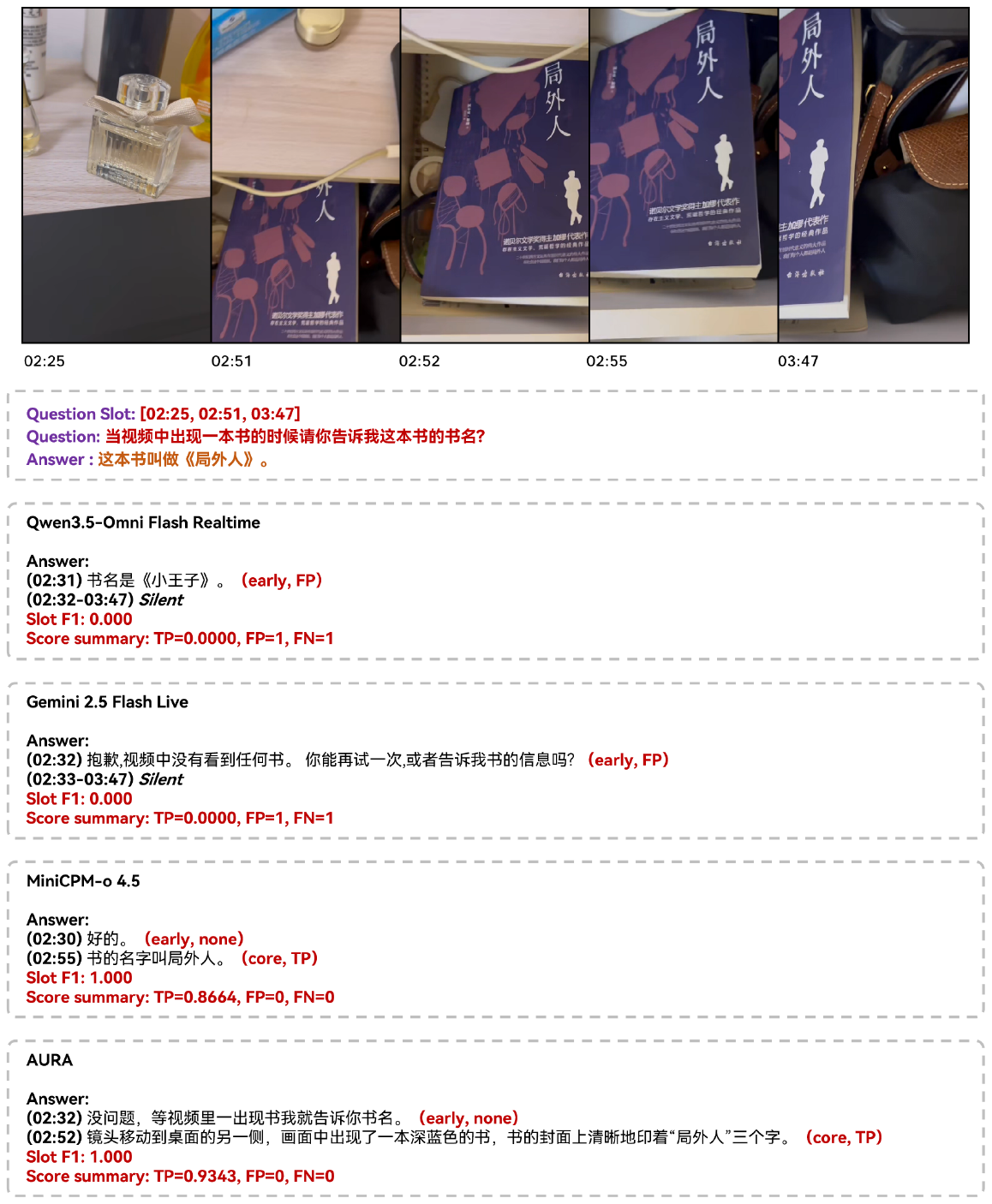}
    \caption{Proactive interaction case. The model must wait until a book appears and then report its title.}
    \label{fig:case_proactive}
\end{figure*}

\begin{figure*}[t]
    \centering
    \includegraphics[width=\linewidth,height=0.9\textheight,keepaspectratio]{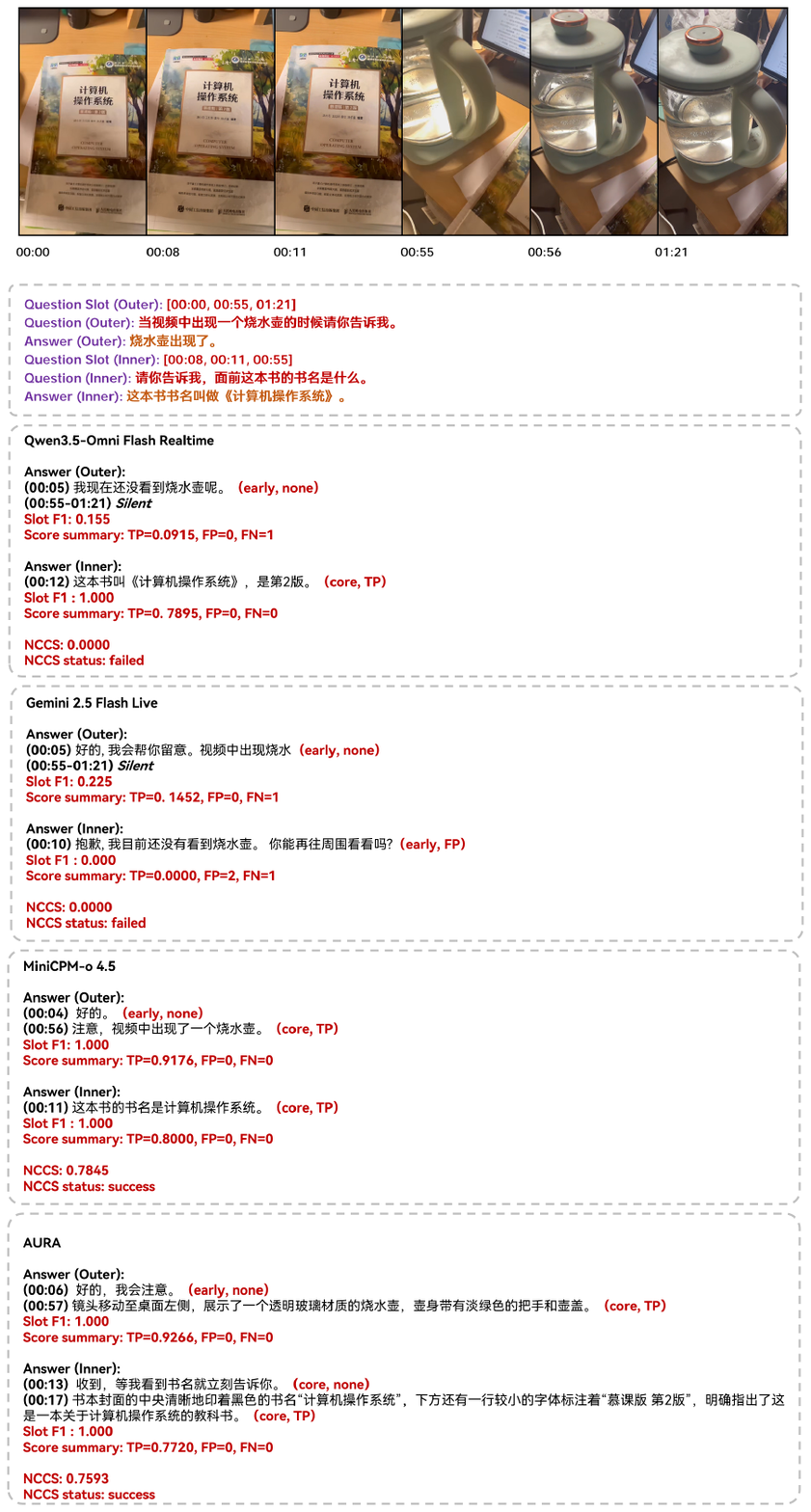}
    \caption{Nested interaction case. The model first monitors for a kettle, then answers an inserted book-title question, and finally should resume the outer monitoring task.}
    \label{fig:case_nested}
\end{figure*}

\begin{figure*}[t]
    \centering
    \includegraphics[width=\linewidth,height=0.9\textheight,keepaspectratio]{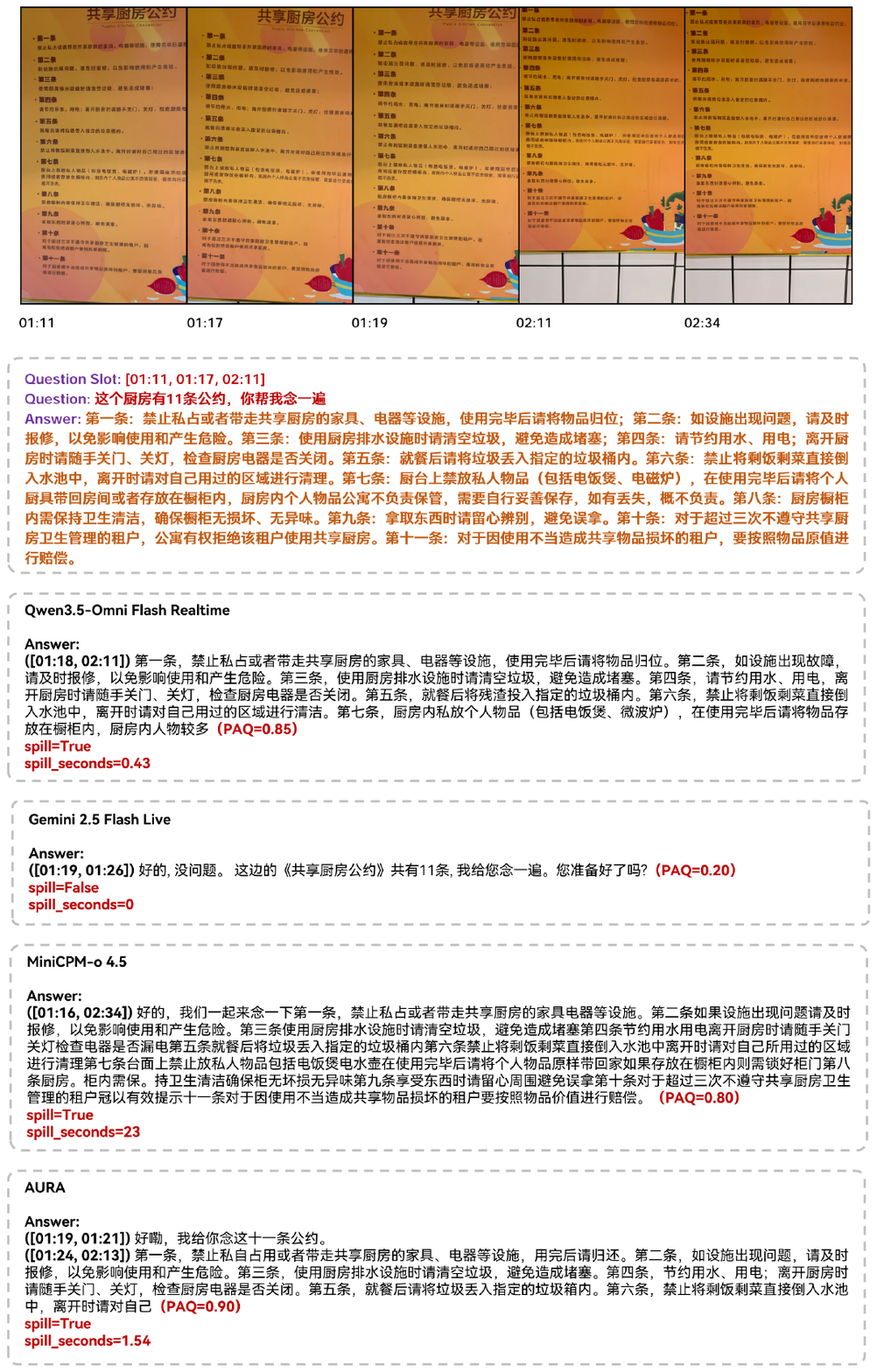}
    \caption{Interruption case. The user asks the model to read public-kitchen rules, but the answer window is truncated by an interruption.}
    \label{fig:case_interruption}
\end{figure*}

\begin{figure*}[t]
    \centering
    \includegraphics[width=\linewidth,height=0.9\textheight,keepaspectratio]{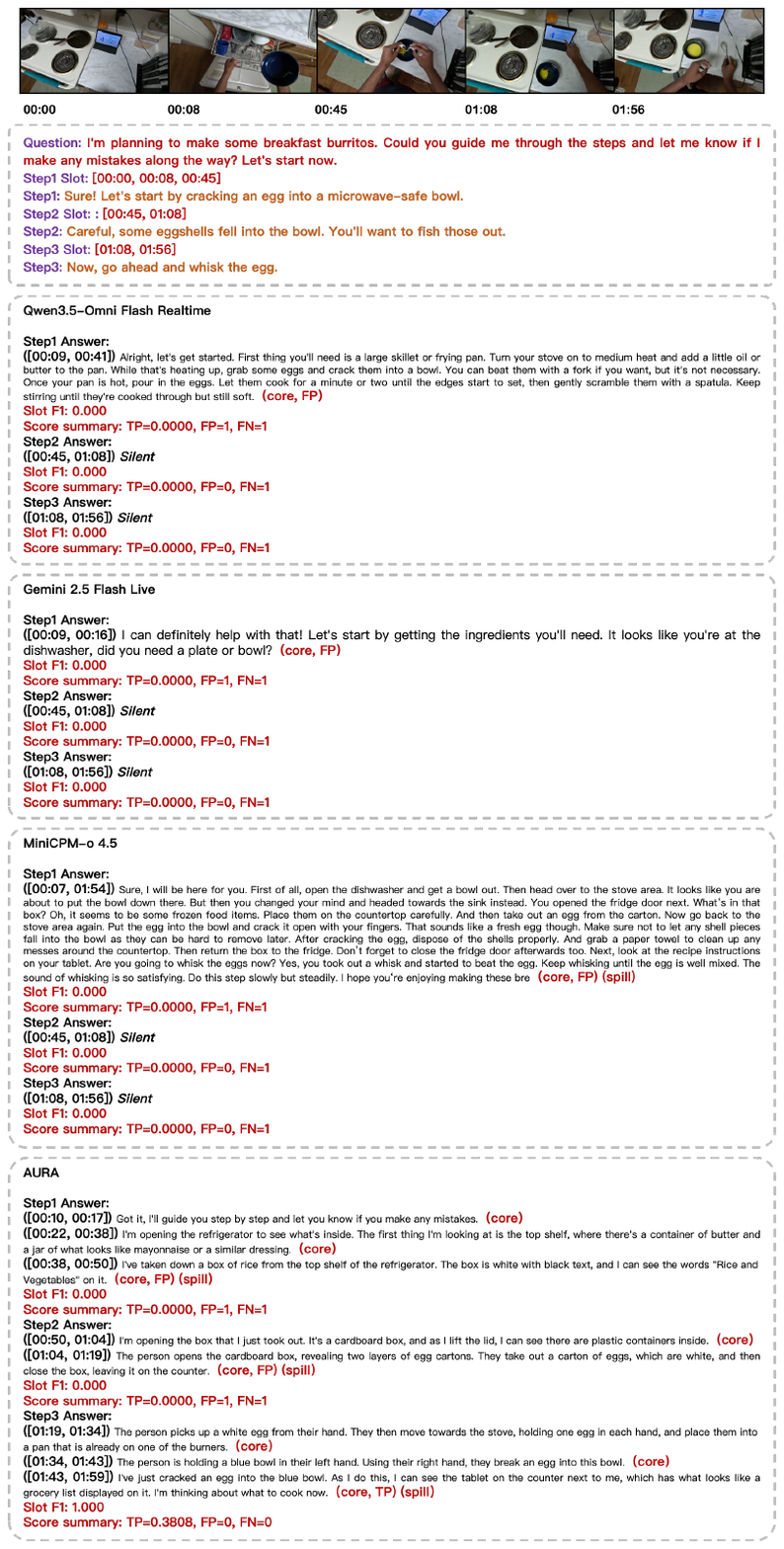}
    \caption{1QnA case. A breakfast-burrito instruction requires multiple temporally grounded responses across a continuous procedure.}
    \label{fig:case_1qna}
\end{figure*}

We provide qualitative examples in Figs.~\ref{fig:case_realtime}--\ref{fig:case_1qna} to illustrate the behaviors behind the aggregate results in Sec.~\ref{sec:exp}. Each example shows sampled video frames, the annotated interaction slot, the reference answer, and model outputs aligned to early/core segments. These cases make two patterns visible: models often possess the local perceptual ability needed to answer a frame-grounded question, but they frequently fail when the interaction requires deciding when to speak, when to wait, when to stop, or when to resume a suspended goal.

\textbf{Annotation convention.}
The TP/FP/FN tags in the case figures denote slot- or stage-level outcomes, not independent per-chunk judgments. When several chunks belong to the same evaluated stage, the tag is placed on the last chunk to summarize the concatenated response judged for that stage. A ``spill'' tag indicates output beyond a hard interruption or slot boundary and may add an FP; 1QnA slots use soft boundaries, so slight carry-over between adjacent steps is tolerated and is not by itself counted as a spill FP. PAQ denotes the Partial Answer Quality score for interrupted slots, measuring the usefulness of the already spoken partial response without requiring completion.

\textbf{Real-time visual question answering.}
In Fig.~\ref{fig:case_realtime}, the user asks for the refrigerator's energy-efficiency level in the slot $[04{:}06,04{:}11,05{:}01]$. The visual evidence is localized: the label becomes readable around the valid-answer time, and the correct response is that the refrigerator is level~1 energy efficient. Gemini and Qwen3.5-Omni produce acceptable core answers, with TP scores of $0.7935$ and $0.7123$, respectively. In contrast, AURA and MiniCPM-o both answer that the refrigerator is level~2 energy efficient, yielding TP scores of $0.0000$ with FP/FN penalties. This case supports the observation in Sec.~\ref{sec:exp} that explicit real-time queries are relatively easier than stateful interactions, but also shows that localized perception can still fail when the model misreads the fine-grained visual attribute.

\textbf{Proactive response timing.}
Fig.~\ref{fig:case_proactive} shows a proactive book-title query: the user asks the assistant to report the title when a book appears, and the correct title is \textit{The Stranger}. AURA waits with an acknowledgment and answers after the book becomes visible, achieving a TP score of $0.9343$. MiniCPM-o behaves similarly and obtains a TP score of $0.8664$. In contrast, Gemini responds in the early stage that no book is visible and asks the user to try again, while Qwen3.5-Omni prematurely guesses \textit{The Little Prince}. Both are counted as early hallucinations and receive FP/FN penalties. The example explains why proactive IA-QTF1 favors MiniCPM-o and AURA in Tab.~\ref{tab:unified_final}: success depends less on recognizing the final object alone and more on suppressing premature answers until the trigger is actually supported by the stream.

\textbf{Nested context switching and resumption.}
The nested case in Fig.~\ref{fig:case_nested} combines an outer monitoring instruction (notify the user when a kettle appears) with an inserted inner query asking for the title of a visible book. MiniCPM-o answers the inner question immediately, then resumes the outer task when the kettle appears, yielding successful NCCS with a score of $0.7845$. AURA also completes both parts with NCCS $0.7593$, although it uses more descriptive wording. Gemini fails because it treats the inner query as if the outer kettle task were still the active question, producing an early response about not seeing a kettle instead of reading the book title. Qwen3.5-Omni answers the inner book-title question correctly, but never resumes the outer monitoring task, so NCCS is zero despite a valid inner answer. This qualitative pattern matches Tab.~\ref{tab:nested}: many models can answer the inserted query locally, but maintaining a suspended outer intent and returning to it remains difficult.

\textbf{Interruption control.}
Fig.~\ref{fig:case_interruption} isolates the full-duplex stopping problem. The model is asked to read eleven public-kitchen rules, but the slot is interrupted at $02{:}11$, so completion is not required; the key behavior is whether generation stops at the boundary, while the partial content quality indicates whether the model has provided useful information before or around the interruption. Gemini stops before the interruption and has no spill, but its output is mostly a preface rather than the requested rules, yielding a low PAQ score of $0.20$. Qwen3.5-Omni and AURA both read useful rule content and continue only slightly beyond the boundary, with PAQ scores of $0.85$ and $0.90$ and spill durations of $0.43$~s and $1.54$~s, respectively. MiniCPM-o also provides substantive rule content (PAQ $0.80$), but continues reading for about $23$~s after interruption, crossing the boundary with a long answer. This case directly supports the interruption diagnostics in Tab.~\ref{tab:interruption}: no spill alone can mask a lack of useful partial content, while high partial quality must still be considered together with conditional spill behavior.

\textbf{Long-horizon 1QnA monitoring.}
The 1QnA example in Fig.~\ref{fig:case_1qna} asks the model to guide a breakfast-burrito procedure and report mistakes. The first valid instruction is to crack an egg into a microwave-safe bowl; later slots include detecting eggshells in the bowl, prompting the user to whisk the egg, and then microwaving while stirring. All models struggle across the first four response slots. Gemini gives an irrelevant dishwasher-related response and then misses later slots. Qwen3.5-Omni answers with a generic skillet-based recipe, revealing unsupported future steps instead of tracking the observed procedure. MiniCPM-o produces a long monologue that rolls multiple future actions into one response, causing spill and losing temporal alignment. AURA is the only model with a nonzero score in the shown slots, but its valid response is delayed and partial, and it still misses the error-correction and next-step guidance. This case illustrates why all models have very low 1QnA IA-QTF1 in Sec.~\ref{sec:exp}: continuous task assistance requires a sequence of small, timely decisions, so one early over-generation or missed event can degrade multiple slots.

Overall, the cases show that OmniInteract penalizes failures that are central to real streaming assistance: guessing before the evidence appears, missing when to respond, forgetting a paused request, and continuing after interruption. They therefore provide qualitative support for the main experimental conclusions: explicit localized queries are comparatively tractable; proactive and nested interactions expose state-management weaknesses; interruption handling varies sharply across models; and long-horizon 1QnA remains the most challenging setting.

%% file: tabs/data_license.tex
\begin{table*}[t]
\centering
\small
\setlength{\tabcolsep}{4pt}
\renewcommand{\arraystretch}{1.08}
\caption{Licenses and access terms for external data sources and data-generation tools used in OmniInteract.}
\label{tab:data_license}
\begin{tabularx}{\textwidth}{p{0.24\textwidth}p{0.25\textwidth}p{0.25\textwidth}X}
\toprule
\textbf{Source} & \textbf{Use in OmniInteract} & \textbf{License / access terms} & \textbf{Notes} \\
\midrule
Qualcomm Interactive Cooking Dataset~\citep{bhattacharyya2026can}
& 1QnA live step-by-step task guidance instances
& Data License Agreement -- Research Use
& The dataset provides manually annotated instructions, feedback messages, and timestamps; its videos are from CaptainCook4D. \\
\midrule
CaptainCook4D~\citep{peddi2024captaincook4d}
& Procedural activity videos and annotations used directly and through the Qualcomm Interactive Cooking Dataset
& Apache License 2.0 for the dataset; MIT License for the public annotation repository
& We follow the license terms of the specific files used. \\
\midrule
EgoPER~\citep{lee2024error}
& 1QnA egocentric procedural error-detection instances
& Available by request under the original dataset owners' access terms
& The official project page requires users to request dataset access from the authors and provide institutional and research-purpose information. \\
\midrule
Qwen3-TTS~\citep{hu2026qwen3}
& Text-to-speech synthesis for initial 1QnA spoken instructions
& Apache License 2.0
& Used only to synthesize the initial instruction audio prepended to the original audio-visual streams. \\
\bottomrule
\end{tabularx}
\vspace{-0.5em}
\end{table*}

%% file: tabs/detail_tab.tex
\begin{table*}[t]
\centering
\small
\setlength{\tabcolsep}{4pt}
\renewcommand{\arraystretch}{0.95}
\caption{Detailed TP/FP/FN breakdown per interaction category. 1Q1A categories are mutually exclusive. ``All Global'' aggregates all 1,430 slots; its FP includes unmatched chunks not attributed to individual categories.}
\label{tab:breakdown}
\vspace{-0.5em}
\begin{tabular}{ll rrr rrr}
\toprule
\textbf{Model} & \textbf{Category} & \textbf{TP} & \textbf{FP} & \textbf{FN} & \textbf{Precision} & \textbf{Recall} & \textbf{IA-QTF1} \\
\midrule
\multirow{5}{*}{AURA}
& Real-time  & 162.42 & 293 & 245 & 0.357 & 0.399 & 0.376 \\
& Proactive  & 90.59  & 95  & 54  & 0.488 & 0.627 & 0.549 \\
& Nested     & 116.77 & 73  & 85  & 0.615 & 0.579 & 0.596 \\
& 1QnA       & 12.52  & 159 & 294 & 0.073 & 0.041 & 0.052 \\
& \textbf{All Global} & \textbf{382.30} & \textbf{666} & \textbf{678} & \textbf{0.365} & \textbf{0.361} & \textbf{0.363} \\
\midrule
\multirow{5}{*}{\shortstack[l]{Gemini 2.5\\Flash Live}}
& Real-time  & 211.32 & 163 & 179 & 0.565 & 0.541 & 0.553 \\
& Proactive  & 16.77  & 73  & 170 & 0.187 & 0.090 & 0.121 \\
& Nested     & 69.06  & 50  & 159 & 0.580 & 0.303 & 0.398 \\
& 1QnA       & 4.97   & 37  & 314 & 0.118 & 0.016 & 0.028 \\
& \textbf{All Global} & \textbf{302.12} & \textbf{328} & \textbf{822} & \textbf{0.480} & \textbf{0.269} & \textbf{0.344} \\
\midrule
\multirow{5}{*}{MiniCPM-o 4.5}
& Real-time  & 150.10 & 326 & 264 & 0.315 & 0.363 & 0.337 \\
& Proactive  & 97.27  & 62  & 64  & 0.611 & 0.603 & 0.607 \\
& Nested     & 119.65 & 72  & 88  & 0.624 & 0.576 & 0.599 \\
& 1QnA       & 2.92   & 70  & 316 & 0.040 & 0.009 & 0.015 \\
& \textbf{All Global} & \textbf{369.94} & \textbf{539} & \textbf{732} & \textbf{0.407} & \textbf{0.336} & \textbf{0.368} \\
\midrule
\multirow{5}{*}{\shortstack[l]{Qwen3.5-Omni\\Flash Realtime}}
& Real-time  & 216.12 & 218 & 174 & 0.498 & 0.554 & 0.524 \\
& Proactive  & 17.23  & 113 & 171 & 0.132 & 0.092 & 0.108 \\
& Nested     & 76.88  & 112 & 140 & 0.407 & 0.354 & 0.379 \\
& 1QnA       & 4.41   & 68  & 314 & 0.061 & 0.014 & 0.023 \\
& \textbf{All Global} & \textbf{314.64} & \textbf{522} & \textbf{799} & \textbf{0.376} & \textbf{0.283} & \textbf{0.323} \\
\bottomrule
\end{tabular}
\end{table*}